\documentclass{article}
\usepackage{ismir,amsmath,cite}
\usepackage[hyphens]{url}
\usepackage{graphicx}
\usepackage{color}
\usepackage{mathtools}
\usepackage{amsfonts}
\usepackage{enumitem}
\usepackage{nccmath}

\newcommand\EatDot[1]{}

\DeclarePairedDelimiterX{\infdivx}[2]{(}{)}{%
  #1\;\delimsize\|\;#2%
}

\title{Learning a Latent Space of Multitrack Measures}

\multauthor
{Ian Simon \hspace{3.15cm} Adam Roberts \hspace{3.0cm} Colin Raffel}
{\bfseries{
Jesse Engel \hspace{2.7cm} Curtis Hawthorne \hspace{2.7cm} Douglas Eck}\\
Google Brain\\
{\tt\small \{iansimon,adarob,craffel,jesseengel,fjord,deck\}@google.com}
}
\def\authorname{Ian Simon, Adam Roberts Colin Raffel, Jesse Engel, Curtis Hawthorne, Douglas Eck}

\begin{document}

\maketitle
\begin{abstract}
Discovering and exploring the underlying structure of multi-instrumental music using learning-based approaches remains an open problem.  We extend the recent MusicVAE model  \cite{roberts2018hierarchical} to represent multitrack polyphonic measures as vectors in a latent space.  Our approach enables several useful operations such as generating plausible measures from scratch, interpolating between measures in a musically meaningful way, and manipulating specific musical attributes.  We also introduce chord conditioning, which allows all of these operations to be performed while keeping harmony fixed, and allows chords to be changed while maintaining musical ``style''.  By generating a sequence of measures over a predefined chord progression, our model can produce music with convincing long-term structure.  We demonstrate that our latent space model makes it possible to intuitively control and generate musical sequences with rich instrumentation (see \url{https://goo.gl/s2N7dV} for generated audio).
\end{abstract}

\section{Introduction}
\label{sec:introduction}

Recent advances in machine learning have made it possible to train generative models which can accurately represent and generate many different types of objects such as images, sketches \cite{ha2017neural}, and piano performances \cite{performance-rnn-2017}, to name a few.  Some of these models learn a {\it latent space}: a lower-dimensional representation that can be mapped to and from the object space.  A major advantage of such latent space models is that many operations that would be difficult to perform in the object space, like morphing between two objects in a semantically meaningful way, become straightforward arithmetic in the latent space. It has even been claimed that latent space models can augment human understanding of the object domain \cite{carter2017using}.  Latent space models have already been trained for several musical concepts including raw waveforms of notes \cite{engel2017neural}, melodies and drum tracks \cite{roberts2018hierarchical}, and playlists \cite{latent-playlists}.  Such models are also frequently used for music recommendations \cite{koren2009matrix}, where both user ``taste'' and song ``style'' are reasoned about in terms of latent vectors.

In this paper, we present a latent space model of individual measures of music with multi-instrument polyphony and dynamics.  One way to think about such objects is as {\it musical textures}; however, we do not model the audio itself but rather use a symbolic representation of the music.  This latent space model allows us to perform a number of intuitive operations:
\begin{itemize}[nolistsep]
\item Sample a measure from the prior distribution to generate novel music from scratch.
\item Interpolate (i.e. slowly morph) between two measures in a semantically meaningful way.
\item Apply attribute transformations to an existing measure, e.g. ``increase note density'' or ``add strings''.
\end{itemize}
The latent space model can also be augmented with additional conditioning variables, which we demonstrate with chords.  Chord conditioning allows us to perform the above operations while holding chords constant or to change chords while keeping musical texture constant.

Even though this model only represents individual measures and thus is incapable of generating long-term structure on its own, combining the latent space with chord conditioning makes it fairly easy to generate music with convincing long-term dependencies; e.g.\ a composer could pick a single point in the latent space and then decode that point (or slowly interpolate between two points) over a desired chord progression.  The contributions of this paper are as follows:
\begin{enumerate}[nolistsep]
    \item An extension of the MusicVAE model \cite{roberts2018hierarchical} to handle up to 8 tracks played by arbitrary MIDI programs.
    \item A novel event-based track representation that handles polyphony, micro-timing, dynamics, and instrument selection.
    \item Introducing chord-conditioning to a latent space model, so that chords and arrangement/orchestration can be controlled independently.
\end{enumerate}

\section{Related Work}

Our work builds on a long history of past efforts at symbolic music generation, plus more recent interest in latent spaces and modeling interplay between instruments.  Algorithmic music generation has been a topic of interest for at least 200 years \cite{kirnberger1767allezeit,menabrea1843sketch,roure2016numbers}.  Prior to the recent neural network renaissance, most systems, such as that of Cope \cite{cope1991computers}, used human-encoded rules, Markov models, or a few other method categories as described by Fern{\'a}ndez and Vico \cite{fernandez2013ai} and Papadopoulos and Wiggins \cite{papadopoulos1999ai} in recent surveys.

The use of neural networks in symbolic music generation has seen a resurgence in interest, as surveyed by Briot et al. \cite{briot2017deep}.  Early work on neural networks for symbolic music generation includes Bharucha and Todd \cite{bharucha1989modeling}, Mozer \cite{mozer1991connectionist}, Chen and Miikkulainen \cite{chen2001creating}, and Eck and Schmidhuber \cite{eck2002finding}.  One of the first effective neural network systems for generating polyphonic music is from Boulanger-Lewandowski et al. \cite{boulanger2012modeling}, who use a recurrent model over a pianoroll representation to generate classical and folk music.  The pianoroll is a fairly standard representation for polyphonic music generation; we instead use an event-based representation closer to the MIDI standard itself.

Some work in music generation has been focused on specific domains.  One such popular domain for polyphonic music generation is Bach chorales: DeepBach \cite{hadjeres2016deepbach}, CoCoNet \cite{huang2017coco}, and BachBot \cite{liang2017automatic} are all different generative models for polyphonic Bach chorales that can respond to user input.  In contrast, our model presented in this paper works simultaneously across multiple Western music styles including classical, jazz, and pop/rock; essentially any music expressible with MIDI notes and program changes is compatible.  Because of its latent space representation, our model is also able to interpolate between these different domains.

Other recent systems for generating polyphonic music include JamBot \cite{brunner2017jambot} which generates chords then notes in a two-step process, DeepJ \cite{mao2018deepj} which generates polyphonic piano music where a user can control several style parameters, a model from Roy et al. \cite{Roy2017SamplingVO} that generates variations on lead sheets, and Song from PI \cite{DBLP:journals/corr/ChuUF16} which generates melody, chord, and drum tracks using a hierarchical recurrent network combined with hand-engineered features.

There are also commercial software systems such as PG Music's Band-in-a-Box \cite{gannon1990band} and Technimo's iReal Pro \cite{irealpro2008} that generate multi-instrumental music over user-specified chord progressions.  However, these products appear to support a limited and fixed number of preset arrangement styles combined with rule-based modifications.

Perhaps the most similar systems to our current work are MusicVAE from Roberts et al. \cite{roberts2018hierarchical} (which we directly extend) and MuseGAN from Dong et al. \cite{dong18aaai} (which builds upon the work of Yang et al. \cite{yang2017midinet}).  MuseGAN \cite{dong18aaai} is based on generative adversarial networks (GANs) and is capable of modeling multiple instruments over multiple bars.  Like our work, the system uses a latent space shared across tracks to handle interdependencies between instruments. However, in MuseGAN the set of instruments is a fixed quintet consisting of bass, drums, guitar, piano, and strings, whereas our system handles arbitrary instrument combinations.  Separately, MuseGAN is focused on accompaniment and generation and is unable to represent or manipulate preexisting music.  Our system can also generate from scratch, but in contrast with MuseGAN, it can also facilitate user-driven manipulation of existing music via the latent space.

The MusicVAE architecture introduced by Roberts et al. \cite{roberts2018hierarchical} is able to learn a latent space of musical sequences using a novel hierarchical decoder that allows it to model long-term structure and multi-instrument sequences.  However, this work applies strict constraints to the sequences to reach its goals.  In order to guarantee a constant number of events per measure, non-drum tracks are limited to monophonic sequences, and all tracks are represented with a single velocity and quantized at the level of 16\textsuperscript{th} notes.  This reduces the challenge of modeling longer-term structure, but at the expense of expressiveness.  Furthermore, while the ``trio'' MusicVAE is capable of modeling three broad instrument classes--melody, bass, and drums--it is limited to exactly three instruments, arbitrarily excluding potentially pivotal voices and disregarding the specific identity of each instrument (e.g. eletric guitar and piano are both considered ``melody'' instruments).  Further, Roberts et al.\ do not consider modeling fine-grained timing and velocity, nor do they develop a method for chord conditioning. Nevertheless, the MusicVAE architecture and its implementation provide a powerful basis for exploring a more expressive and complete multitrack latent space, and thus we position our work as an extension of it.

\section{Measure Representation}
\label{sec:measure_representation}

\begin{figure}
    \centering
    \includegraphics[width=0.4\textwidth]{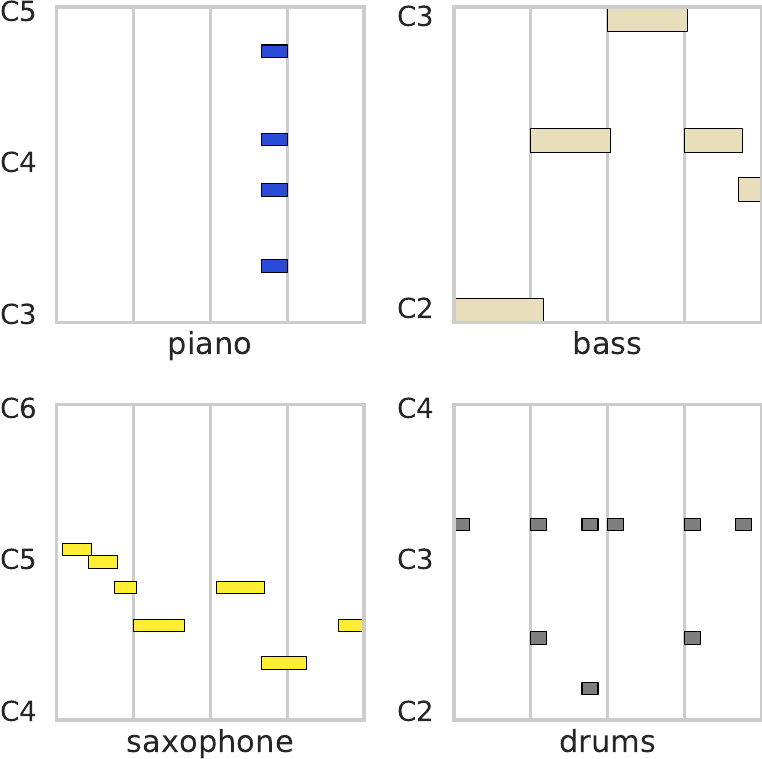}
    \caption{A single measure consisting of 4 tracks, shown as separate pianorolls.  Each track is represented as a MIDI-like sequence of {\it note-on}, {\it note-off}, {\it time-shift}, and {\it velocity-change} events, with a single {\it program-select} event at the beginning and an {\it end-track} event at the end.  For all other figures in this paper, multiple tracks are combined into a single pianoroll, color-coded by instrument family.}
    \label{fig:tracks}
\end{figure}


We model measures with up to 8 tracks (see Figure \ref{fig:tracks} for an example with 4 tracks).  Each track consists of a single ``instrument'' as extracted by \texttt{pretty\_midi} \cite{pretty_midi}.  A track is represented as a MIDI-like sequence of events from an extension of the vocabulary used by Simon and Oore  \cite{performance-rnn-2017} to handle metric timing and choice of instrument:
\begin{itemize}[nolistsep]
\item 128 {\bf note-on} events, one for each MIDI pitch.
\item 128 {\bf note-off} events, one for each MIDI pitch.
\item 8 {\bf velocity-change} events, MIDI velocity quantized into 8 bins.  These events set the velocity for subsequent note-on events.
\item 96 {\bf time-shift} events that shift the current time forward by the corresponding number of quantized time steps, where 24 steps is the length of a quarter note.
\item 129 {\bf program-select} events (128 programs plus drums) that set the MIDI program number at the beginning of each track.
\item A single {\bf end-track} event, used to mark the end of each track.  For measures with fewer than 8 tracks, missing tracks consist solely of the end-track event.
\end{itemize}
For simplicity we only include measures with exactly 96 quantized time steps (4 quarter notes), as this is the most frequent measure size in our dataset.

\section{Model}

\begin{figure*}[t]
    \centering
    \includegraphics[width=0.85\textwidth]{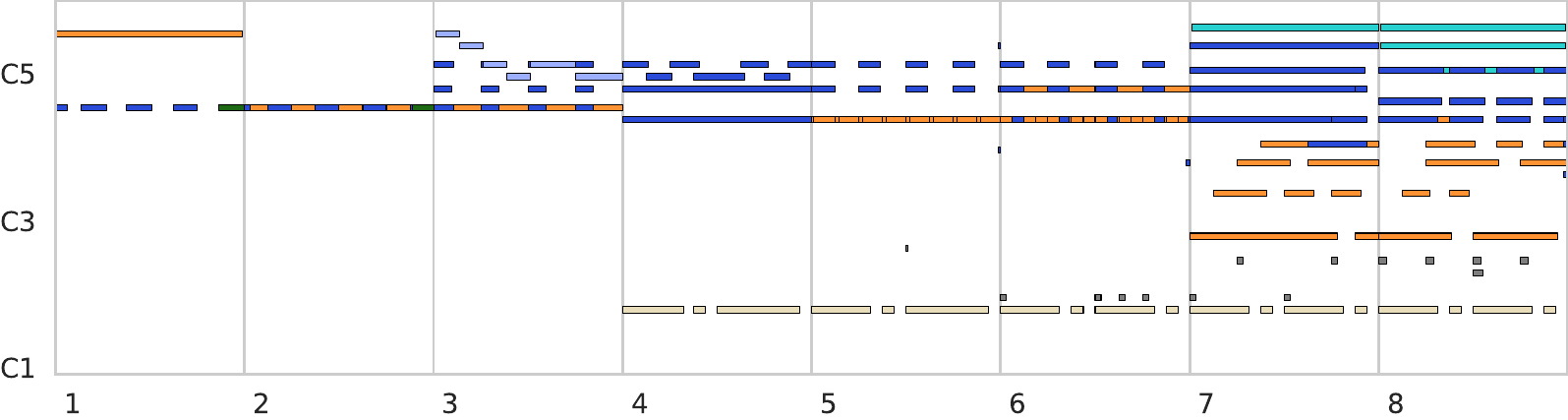}
    \caption{An interpolation between two measures generated by our model.}
    \label{fig:interpolations}
\end{figure*}

\begin{figure*}[t]
    \centering
    \includegraphics[width=0.85\textwidth]{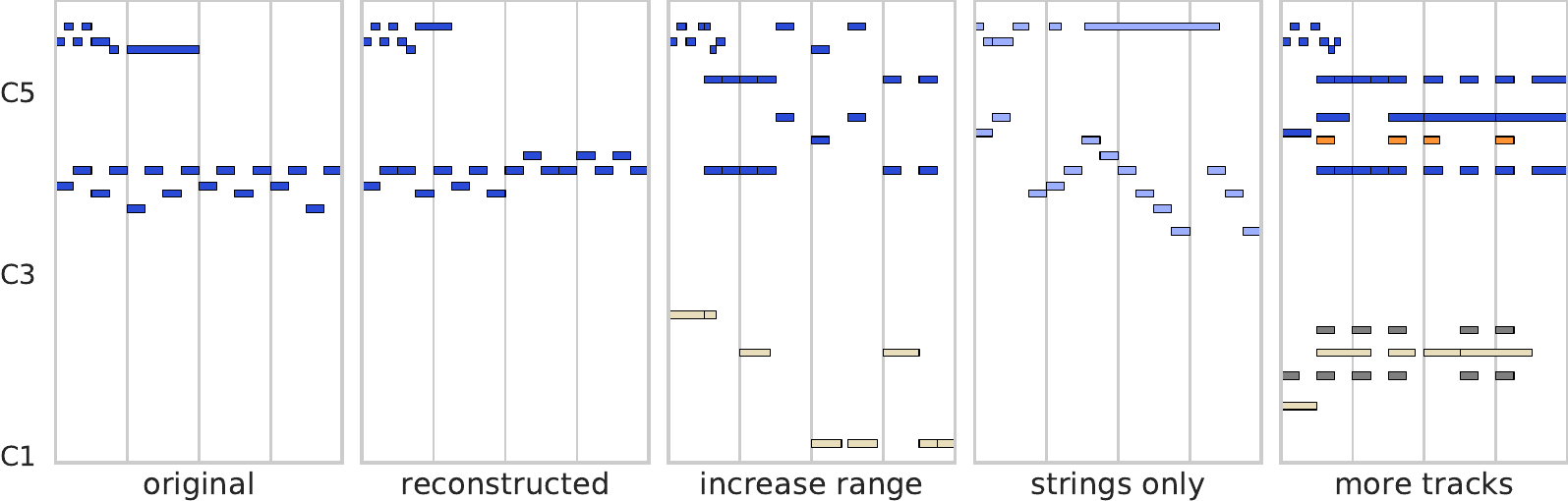}
    \caption{Multiple transformations to a single measure via attribute vector arithmetic.  On the left is the original measure, followed by its reconstruction from the latent space.  After that are three transformations: increasing the pitch range, using only string instruments, and using more tracks.}
    \label{fig:attributes}
\end{figure*}

\begin{figure*}[t]
    \centering
    \includegraphics[width=0.85\textwidth]{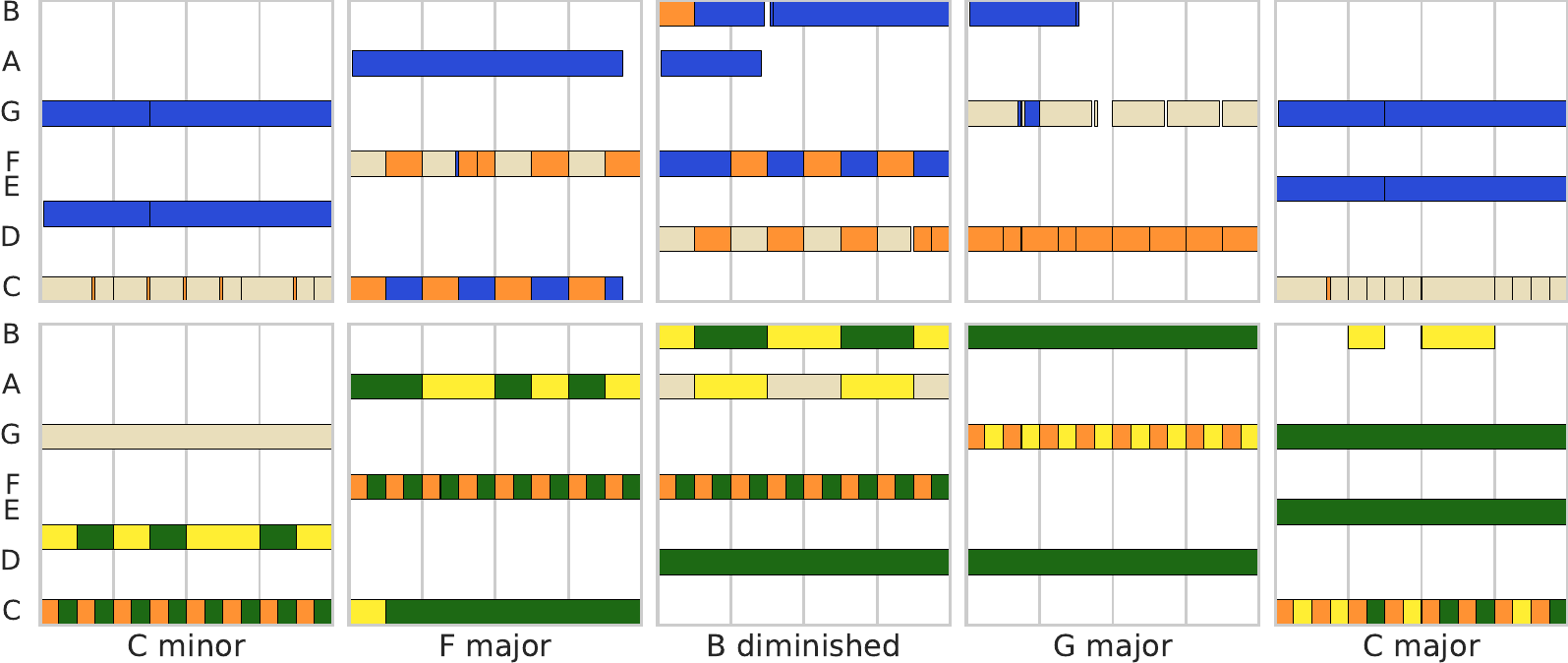}
    \caption{Two points in the latent space, each decoded over five different chords.  Drums and pitch octave information have been removed from the pianorolls to show that the model is respecting the chord conditioning.}
    \label{fig:chords}
\end{figure*}

\begin{figure*}[t]
    \centering
    \includegraphics[width=0.85\textwidth]{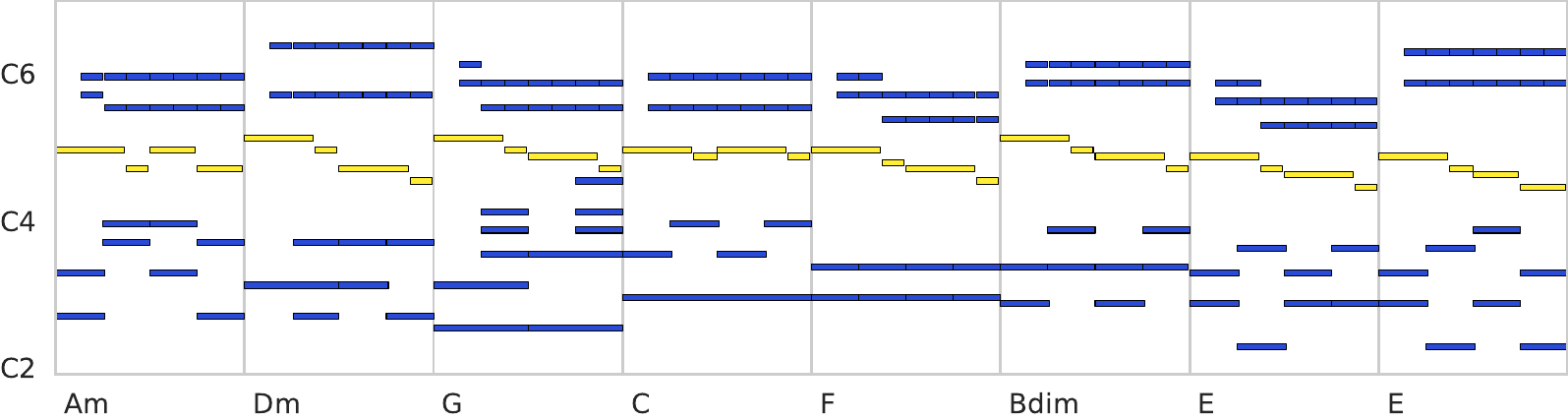}
    \caption{A single latent point decoded over a chord progression.}
    \label{fig:chord_progression}
\end{figure*}

Our model is a variational autoencoder (VAE) \cite{kingma2014stochastic} over hierarchical sequences, extending the architecture of MusicVAE \cite{roberts2018hierarchical} to handle variable numbers of tracks and events per track.  The top level of the decoder hierarchy produces {\it track embeddings} from a single latent code and the bottom level produces the sequence for each track, including the choice of instrument as a MIDI program number. The encoder (which is also hierarchical) works in reverse, first converting each track to a single vector then mapping this sequence of track vectors to the latent space.  We achieve additional control over the output of the model by adding chord conditioning to both the encoder and decoder (see Section \ref{subsec:conditioning} for details).  In the following subsections, we provide a brief overview of our model; for a more in-depth description of our baseline, see \cite{roberts2018hierarchical}.


\subsection{Variational Autoencoders}
An {\it autoencoder} is a model that learns to ``compress'' or encode objects to a lower-dimensional latent space and then decode these latent representations back into the original objects.  Autoencoders are typically trained to optimize a {\it reconstruction loss} which measures how close the reconstructed version is to the original object.  This encourages the model to produce a compressed representation that captures the important variation among the objects.

The variational autoencoder extends the basic autoencoder in that it considers the latent representation $z$ to be a {\it random variable} drawn from a {\it prior} distribution $p(z)$, usually a multivariate Gaussian with diagonal covariance. The encoder approximates the {\it posterior} distribution $p(z \mid x)$, while the decoder models the {\it likelihood} $p(x \mid z)$.  The VAE is thus a generative model with the following generation process: generate a latent vector $z$ from the prior distribution by sampling $z \sim p(z)$, then use the decoder to sample an output using the sampled $z$ by $x \sim p(x \mid z)$.

In a variational autoencoder, the encoder and decoder are typically neural networks $q_\lambda(z \mid x)$ and $p_\theta(x \mid z)$ parameterized by $\theta$ and $\lambda$, respectively.  In our case where we are dealing with sequences, both the encoder and decoder are hierarchical LSTM \cite{hochreiter1997long} models as in MusicVAE.

This architecture has two key benefits: First, it is a {\it latent variable model}; we can sample new measures and we can map existing measures to the latent space and transform them in various ways.  Second, it is {\it hierarchical}; individual tracks are independent conditional on their embeddings.  The use of track embeddings gives the model a way to represent complex dependencies between tracks, so that tracks will ``fit together" when generated.

\subsection{Loss Function}

The VAE model optimizes a loss function that is the difference of two terms: the reconstruction loss, and the Kullback-Leibler (KL) divergence loss:
\begin{equation}
    \mathbb{E}\left[\log p_\theta(x \mid z) \right] - D_\mathrm{KL}\infdivx{q_\lambda(z \mid x)}{p(z)}
\end{equation}
The reconstruction loss term $\mathbb{E}\left[\log p_\theta(x \mid z) \right]$ maximizes the log-likelihood of the training data.  In our model, the reconstruction loss is computed as the sum of the cross entropy between the predicted output distribution and the ground truth value over all events on all tracks in a given sequence.  The KL divergence loss term \mbox{$D_\mathrm{KL}\infdivx{q_\lambda(z \mid x)}{p(z)}$} encourages $q_\lambda(z \mid x)$ (the distribution produced by the encoder) to be close to $p(z)$, the unit Gaussian prior.

One implication of optimizing a combination of loss terms is that there's a core tradeoff between two desires:
\begin{enumerate}[nolistsep]
    \item {\it reconstruction}: The model should be able to faithfully represent measures from the training set in the latent space, such that these measures can be reproduced from their latent code.
    \item {\it sampling}: The prior should be enforced; i.e. the posterior distributions for encoded measures from the training set should be close to the prior.  This ensures that measures sampled from the prior are plausible.
\end{enumerate}
We control this tradeoff by using the ``free bits'' method of Kingma et al. \cite{kingma2016improved}; the KL loss term is allowed a budget $\tau$ bits of entropy per training example before it begins accruing loss.  Increasing $\tau$ therefore improves reconstruction fidelity with the drawback of less realistic samples and semantically meaningless interpolation.  We found $\tau = 64$ produced a good trade-off between reconstruction and sampling/interpolation in most cases.  The only exception was in the attribute vector experiments described in Section \ref{subsec:attribute}, for which we found better results by using $\tau = 256$.  Perhaps surprisingly, providing chords (see Section \ref{subsec:conditioning}) had little effect on reconstruction accuracy (for a given $\tau$) even though samples from the chord-conditioned model do respect the chord conditioning.

\subsection{Architecture}
In this section, we give a brief overview of our model's architecture; for specific details refer to \cite{roberts2018hierarchical} and our public source code\footnote{\url{https://github.com/tensorflow/magenta}}.  Our encoder consists of two levels of bidirectional LSTMs.  The first level independently consumes each of the 8 track event streams, concatenating the final outputs from the forward and backward directions to produce 8 track embeddings.  The second level consumes these 8 embeddings, outputting a single embedding that is the concatenation of final states of the forward and backward directions.  This embedding is then fed through two fully-connected layers to produce the $\mu$ and, after a softplus activation, the $\sigma$ parameter for the autoencoder's latent distribution.  A latent vector is then sampled from a multivariate Gaussian distribution with a diagonal covariance, parameterized by $\mu$ and $\sigma$.

The decoder is made up of two levels of unidirectional LSTMs.  The first level (called the ``conductor'' by Roberts et al.)\ is initialized by setting its state to be the result of passing the latent vector though a linear layer with a $\tanh$ activation. This conductor LSTM is then run for 8 steps with a null input, outputting 8 track embeddings.  For each of the 8 tracks, the lower-level LSTM is initialized in the same manner as the conductor using one of the 8 track embeddings. The initial input to the LSTM for each track is a zero vector concatenated with the track embedding, and subsequent inputs are the one-hot representation of the previous event concatenated with the track embedding.  The outputs of the LSTM are then passed through a final softmax layer over the event vocabulary.

\subsection{Conditioning}
\label{subsec:conditioning}

While the latent space allows for the manipulation of individual attributes, it is sometimes difficult to avoid introducing side effects on correlated attributes. By conditioning both the encoder and decoder on chords, we encourage the model to ``factor out'' chord information from the latent representation.  This allows us to control chords and other attributes independently, which supports both holding the ``arrangement'' of the sequence constant while changing the underlying chord progression and holding the chord progression constant while changing the arrangement.

We encode a chord as a one-hot vector over 49 chord types: major, minor, augmented, and diminished triads for all 12 pitch classes, plus a ``no-chord'' value.  This chord vector is appended to the model input at each encoding and decoding step and can vary between steps; as such we are able to model harmonic changes within a single measure.

\subsection{Training}

The model is trained with the Adam optimizer \cite{DBLP:journals/corr/KingmaB14} using a batch size of 256. We anneal the learning rate from \mbox{1e-3} to \mbox{1e-5} with exponential decay rate 0.9999, for 100,000 gradient update steps.  We use {\it teacher forcing} and feed the ground truth output value back to the model (instead of using its own output) at each sequence step during training.

For both levels of the model hierarchy (measure-tracks and track-events), we use a bidirectional LSTM encoder with 1024 nodes and a forward LSTM decoder with 3 layers of 512 nodes each. Our latent space has dimension 512.

\subsection{Inference}
\label{subsec:inference}

Many strategies exist for producing a single output sequence from the softmax distributions produced by the LSTM outputs, including beam search and sampling autoregressively with a {\it temperature} parameter that controls the uniformity of the distribution.  In all examples in this paper, we sample autoregressively with a temperature of 0.2 until an end token is returned.

\subsection{Dataset}
Our models are all trained on the Lakh MIDI Dataset \cite{raffel2016learning}, a collection of 176,581 MIDI files scraped from the web.  The dataset is preprocessed as follows:

The dataset is first split into measures, and measures with length different from 4 quarter notes are discarded. Tracks are then extracted from each measure using the \texttt{pretty\_midi} Python library; each track consists of notes with a single MIDI program number (or drums), though multiple tracks may use the same program number.  Measures with fewer than 2 or more than 8 tracks are discarded.  The tracks are then sorted by increasing program number, with drums at the end.  Measures where any one track has more than 64 events (from the vocabulary in Section \ref{sec:measure_representation}) are discarded.

Finally, the measures are deduped, resulting in a training set of 4,092,681 examples. During training, each measure is augmented by tranposing up or down within a minor third by an amount chosen uniformly at random; notes falling outside the valid MIDI pitch range are dropped.  We perform this data augmentation step as the key distribution in the training data is far from uniform; around 50\% of the data set is in C major or A minor.

\subsubsection{Chord Inference}
When conditioning on chords, we would ideally train using ground-truth labels.  Since MIDI files do not typically contain such labels \cite{raffel2016extracting}, we automatically infer chord labels using a heuristic process.

First, each MIDI file is split into segments with a consistent tempo and time signature.  For each segment, we infer chords at a frequency of 2 per measure using the Viterbi algorithm \cite{viterbi1967error} over a heuristically-defined probability distribution; as a byproduct we also infer the time-varying key of the sequence.  This algorithm takes time quadratic in the number of measures, so for efficiency we discard MIDI segments longer than 500 measures.

We infer 8 different chord types (major, minor, augmented, diminished, dominant-seventh, major-seventh, minor-seventh, and half-diminished) rooted at each of the 12 pitch classes plus a single ``no-chord'' designation, for a total of 97 chord classes.  After chord inference is complete, the 8 chord types are projected down to the 4 triad types (49 total classes) used as model input.

Our chord inference computes the maximum-likelihood chord and key sequence over the following probability distribution on keys, chords, and notes:
\begin{equation}
    p(h, y) = p(h_0) p(y_0 \mid h_0) \prod_{t=1}^n p(h_t \mid h_{t-1}) p(y_t \mid h_t)
\end{equation}
where $h$ is the ``harmony'' sequence (key and chord at each step) and $y$ is a sequence of unit-normalized pitch class vectors over the duration-weighted notes at each step.

For simplicity this heuristic approach was designed to minimize the number of parameters while penalizing key changes, chord changes, and key/chord/note pitch mismatches.  Besides the chord change frequency of 2 per bar we use 4 other parameters:
\begin{itemize}[nolistsep]
\item $\gamma = 0.5$, the probability of a chord change
\item $\rho = 0.001$, the probability of a key change
\item $\psi = 0.01$, the probability that a chord note will be drawn from outside the current key
\item $\kappa = 100$, the ``concentration'' of the pitch class distribution under a chord; lower values are more forgiving of pitch mismatches
\end{itemize}
We define the harmony transition distribution as follows:
\begin{equation}
    \medmath{p(h_t \mid h_{t-1}) =  \begin{cases}
    (1 - \gamma) (1 - \rho) & \textrm{if no change} \\
    \gamma (1 - \rho) g(h_t, h_{t-1}) & \textrm{if chord change} \\ 
    \frac{\rho}{11}f(h_t) & \textrm{if key change}
    \end{cases}}
\end{equation}
where $f(h_t)$ is a binomial distribution on the number of chord pitches belonging to the key, and
\begin{equation}
    g(h_t, h_{t-1}) = f(h_t) + \frac{f(h_{t-1})}{48}
\end{equation}
(11 is the number of keys minus the current key, and 48 is the number of chords minus the current chord.)  The note observation distribution is defined as:
\begin{equation}
    p(y_t \mid h_t) \sim \kappa \times \left(y_t \cdot c(h_t)\right)
\end{equation}
where $c(h_t)$ is a unit-normalized vector representing the (uniformly-weighted) pitch classes in the chord for $h_t$.


More complex MIDI-to-chord techniques exist in the literature \cite{scholz2008cochonut, wang2012musical, DBLP:conf/ismir/MasadaB17}, but we find our heuristic approach satisfactory for model conditioning even though it ignores many relevant cues and likely makes basic errors.  For example, even though our heuristic chord inference does not use the fact that the chord root is often played by the bass, the trained VAE model learns this pattern and usually generates bass parts that play the root.

\section{Latent Space Manipulations}

In this section we demonstrate several types of musical manipulations that can be performed via the latent space.  Examples of all of these manipulations can be heard at \url{https://goo.gl/s2N7dV}.

\subsection{Sampling}
Most straightforwardly, we can sample from the model.  By sampling latent codes from the prior distribution and then feeding them through the decoder, we obtain new measures of multitrack music.  Because our model uses a flexible representation and is trained on a large corpus, the samples it generates can be quite diverse.

\subsection{Interpolations}

As demonstrated by Roberts et al. \cite{roberts2018hierarchical}, a latent space can be used to interpolate between two musical sequences in a more semantically meaningful way compared to naively blending the notes together. Given two measures $x_0$ and $x_1$, we can interpolate between them by applying the encoder to obtain latent codes $z_0$ and $z_1$, then for any \mbox{$0 \leq \alpha \leq 1$} constructing $z_\alpha$ using spherical linear interpolation \cite{white2016sampling}, as most of the probability mass of the Gaussian prior lies very close to the unit hypersphere.  We then decode $z_\alpha$ into $x_\alpha$ to obtain the interpolated measure.  Figure \ref{fig:interpolations} shows an 8-step interpolation between two measures constructed in the above manner.

\subsection{Attribute Vector Arithmetic}
\label{subsec:attribute}
Our latent space also makes it fairly straightforward to apply basic manipulations to a sequence.  Given a particular attribute (e.g. note density), we can compute the difference between the mean latent vectors of the set of examples that have the attribute and the set of examples that do not to get an {\it attribute vector}.  Then, given an example sequence that does not have the attribute, we can add the attribute by a) encoding the sequence, b) adding the attribute vector to the latent code, and c) decoding the translated latent code.  As observed by Carter and Nielsen \cite{carter2017using}, this is a rather primitive way to learn such an attribute transformation, but often works in practice.

Figure \ref{fig:attributes} shows several attribute transformations to an example measure: increasing the pitch range, using only string instruments, and using more instruments.  Note that none of these transformations is performed in a straightforwardly mechanical way; indeed, there is often no mechanical way to perform an operation like ``add more tracks''. 

\subsection{Chord Conditioning}
\label{subsec:chord_conditioning}

Figure \ref{fig:chords} shows the same latent vectors decoded under several different chords. Notice that for a given latent vector, the instrumental choice and rhythmic pattern remain fairly consistent, while the harmony changes. This allows us to concatenate multiple measures generated from a single latent vector to create a coherent multi-measure sequence.  We find that this technique approximately matches the playing style of much popular modern music, where players shift a consistent rhythmic pattern--a ``groove''--and modulate it over different repeating chord progressions.  The model also naturally ``vamps'' by introducing small musically-related variations from the same latent vector and chord due to the autoregressive RNN sampling procedure.  Figure \ref{fig:chord_progression} shows such a sequence, which can be listened to at \url{https://goo.gl/s2N7dV} along with other sequences generated in similar fashion.


\section{Conclusion}
We have shown how to train and apply a latent space model over measures of symbolic music with multiple polyphonic instruments.  We believe that ours is the first model capable of generating full multitrack polyphonic sequences with arbitrary instrumentation.  On top of this, many natural musical operations are enabled by our latent space representation including interpolation, attribute manipulation, and (with side information) chord conditioning.  Our source code is available at \url{https://github.com/tensorflow/magenta}.

\section{Acknowledgements}
We would like to thank Anna Huang and Erich Elsen for helpful reviews on drafts of the paper.  SoundFont used in our audio examples created by John Nebauer.

\bibliography{references}

%
%
%
%

\end{document}